%% file: main.tex
\documentclass[10pt,twocolumn,letterpaper]{article}

\usepackage[pagenumbers]{cvpr} 

\usepackage{graphicx}
\usepackage{amsmath}
\usepackage{amssymb}
\usepackage{booktabs}

\usepackage{multirow}
\graphicspath{{figs/}}
\usepackage[dvipsnames]{xcolor}
\usepackage{comment}
\usepackage{booktabs}
\usepackage{caption}
\usepackage{subcaption}
\input{math_commands}

\usepackage{array}
\usepackage{wrapfig}

\usepackage[pagebackref,breaklinks,colorlinks,bookmarks=false]{hyperref}

\usepackage[capitalize]{cleveref}
\crefname{section}{Sec.}{Secs.}
\Crefname{section}{Section}{Sections}
\Crefname{table}{Table}{Tables}
\crefname{table}{Tab.}{Tabs.}

\begin{document}

\title{Uncertainty-based Cross-Modal Retrieval with Probabilistic Representations}

\author{Leila Pishdad$\mathbf{^{1}}$
\quad
Ran Zhang$\mathbf{^{1}}$
\quad
Konstantinos G. Derpanis$\mathbf{^{1,2}}$
\quad
Allan Jepson$\mathbf{^{1,3}}$\\
\quad
Afsaneh Fazly$\mathbf{^{1}}$\\
{\large $\mathbf{^1}$Samsung AI Centre Toronto, $\mathbf{^2}$York University,  $\mathbf{^3}$University of Toronto}\\
{\tt\small leila.pishdad@mail.mcgill.ca },
{\tt\small kosta@yorku.ca},
{\tt\small \{ran.zhang, allan.jepson, a.fazly\}@samsung.com}
}

\maketitle

\begin{abstract}

\input{abstract}
\end{abstract}

\section{Introduction}
\label{sec:intro}
\input{introduction}

\section{Related work}
\label{sec:relwork}
\input{relwork}
\section{Learning probabilistic representations}
\label{sec:model}
\input{model}

\section{Empirical evaluation}\label{sec:Empirical}
\label{sec:results}
\input{results}

\section{Conclusions}
\label{sec:conclusion}
\input{conclusion}

\clearpage

{\small
\bibliographystyle{ieee_fullname}
\bibliography{bib_definitions,prob_rep}
}

\clearpage

\section{Appendix}
\label{sec:sup}
\input{supplemental}

\end{document}

%% file: math_commands.tex
\usepackage{amsmath,amsfonts,bm}

\def\eqref#1{equation~\ref{#1}}

\def\1{\bm{1}}

\def\vmu{{\bm{\mu}}}
\def\vtheta{{\bm{\theta}}}
\def\va{{\bm{a}}}

\def\vsigma{{\bm{\sigma}}}

\DeclareMathAlphabet{\mathsfit}{\encodingdefault}{\sfdefault}{m}{sl}
\SetMathAlphabet{\mathsfit}{bold}{\encodingdefault}{\sfdefault}{bx}{n}

\def\gN{{\mathcal{N}}}

\newcommand{\R}{\mathbb{R}}



%% file: abstract.tex
Probabilistic embeddings have proven useful for capturing polysemous word meanings, as well as ambiguity in image matching. 
In this paper, we study the advantages of probabilistic embeddings in a cross-modal setting (i.e., text and images), and propose a simple approach that replaces the standard vector point embeddings in extant image--text matching models with probabilistic distributions that are parametrically learned.
Our guiding hypothesis is that the uncertainty encoded in the probabilistic embeddings captures the cross-modal ambiguity in the input instances, and that it is through capturing this uncertainty that the probabilistic models can perform better at downstream tasks, such as image-to-text or text-to-image retrieval.
Through extensive experiments on standard and new benchmarks, we show a consistent advantage for probabilistic representations in cross-modal retrieval, and validate the ability of our embeddings to capture uncertainty.

%% file: introduction.tex
In this paper, we consider the challenge of cross-modal retrieval with focus on relating image and text modalities.  The general aim is to learn
a shared representational space, where related instances from multiple modalities are mapped closely together in the joint space. Given the mappings and a query element from one domain, retrieval amounts to finding the nearest (i.e, most relevant) elements in a database in the other domain. 
Most cross-modal retrieval methods (e.g., \cite{Faghri2017,Li2019}) learn mappings of each data item (e.g., an image--caption pair) to a vector representing a point in a $k$-dimensional space. A major issue with these
point embeddings is their expressiveness in terms of the information they can capture about the inputs. For language, a clear drawback of point representations is their inability to handle polysemy, i.e., words or phrases having multiple meanings~\cite{Vilnis2014}.
For images, uncertainty can be introduced by the image formation process (e.g., occlusion)~\cite{KendallG17}. 
For image--text matching, an image can be described in many ways,
and a caption can describe many different images, resulting in some degree of cross-modal ambiguity. 
Point embeddings lack the ability to represent the uncertainties in cross-modal associations and skew to learning an average of the cross-modal information by
regressing to the mean embedding.




In this work, we forgo point mappings and instead relate inputs to a region of the shared representation space. 
The inputs from each modality are modeled 
as probability distributions (i.e., probabilistic embeddings) in a common embedding space.  In particular, the inputs are mapped to Gaussian distributions, where the distribution spread
in the joint embedding space captures the mapping's uncertainty (or ambiguity).
While 
few works have explored similar embeddings in a single modality (i.e., text~\cite{Vilnis2014} or images \cite{Oh2018}) their extension in the cross-modal setting is underexplored with only a single concurrent study~\cite{Chun2021} conducted to date. \newline


\noindent{\bf Contributions.} 
We introduce a simple and general approach to learning probabilistic embeddings (instead of point embeddings) in an image--text matching model, that can explicitly handle and express the inherent uncertainty in establishing cross-modal associations.
Through a comprehensive empirical study of several extant retrieval models and benchmarks, we show a consistent advantage of learning probabilistic embeddings for cross-modal retrieval over 
their point-based counterparts.
Moreover, we show (via a controlled experiment) that a per-instance measure of uncertainty actually captures the cross-modal ambiguity, which we attribute as a key factor for the consistently high performance of the probabilistic models.
We view our contribution as complementary to other innovations in the cross-modal retrieval space (e.g., embedding architectures).
To avoid conflating retrieval performance increases due to the representation
(i.e., point vs.\ probabilistic) with differences in the loss function and/or the backbone architecture, we use standard retrieval methods \cite{Faghri2017,Li2019} in our evaluation and simply replace their terminal point mappings with our probabilistic ones. 
Empirically, our retrieval results exceed or are competitive to 
the adapted baseline point embedding networks, while also 
providing the added benefit of a diagnostic uncertainty measure.  In comparison to the lone recent probabilistic method to cross-modal retrieval \cite{Chun2021}, we demonstrate consistent superior performance 
under all performance metrics.

%% file: relwork.tex


\paragraph{\textbf One-to-one mappings.} For image-based retrieval, most methods are based on (handcrafted or learned) point-based representations of the imagery
\cite{sift,bow,vlad,netvlad,fisher}.  Similarly, standard cross-modal retrieval methods rely on
(independent) one-to-one mappings that take elements from their respective domains (e.g., images or text) as a whole and map them to a shared embedding space, where similarities can be established.  
A popular approach to learn these mappings is based on a 
ranking loss (e.g., \cite{FromeCSBDRM13,Faghri2017,Shi2018,Vendrov2015}), where related elements are encouraged to be closer than unrelated ones.
Alternative models aid learning of the shared embedding space through encoding intra-modality similarities with additional auxiliary losses \cite{Zheng2020}
or leveraging adversarial learning \cite{Chen2021adv}.
As described in Sec.~\ref{sec:intro}, one-to-one mappings cannot handle ambiguities (e.g., polysemy) in the inputs. 
We learn probabilistic embeddings that explicitly encode uncertainty through mapping each input instance to a region instead of a point in a shared representation space. 

\paragraph{\textbf Many-to-many mappings.} To better handle the inherent ambiguities in multimodal inputs, recent methods
\cite{NamHK17,Lee2018,Lu2019,Li2019,Song2019,WangLLSYWS19,LuGRPL20,WeiZLZW20,Wu2019b,ZhangLZL20}
first generate multiple embeddings that capture parts of the input in each domain and then use a module 
to selectively combine the various parts.   
For instance, the model of \cite{Song2019} takes images and sentences, and first generates a set of image regions and word embeddings.
These part embeddings are then combined \emph{independently} in each domain using multi-head self-attention to generate multiple diverse representations for matching.  
Alternatively, a plethora of recent methods 
integrate cross-attention operating \emph{jointly} on image region and word embeddings \cite{HuangWW17,NamHK17,Lee2018,Lu2019,WangLLSYWS19,LuGRPL20,WeiZLZW20,ZhangLZL20,ZhangDG20}, i.e., mixing information across domains.  
While these latter methods are the state of the art in cross-modal retrieval, they require excessive
computation 
that renders them inapplicable for real-world deployment:
for each query, cross-attention is applied to each element in the search database.  
Our focus is on
joint embeddings computed independently in each domain to facilitate large-scale, efficient search. 


\paragraph{\textbf Probabilistic embeddings.} 
Density-based or probabilistic embeddings have been previously used to capture words and their associated
uncertainties in meanings \cite{Vilnis2014,Mukherjee2016,Athiwaratkun2017,Athiwaratkun2018a,Athiwaratkun2018b}.
Our method extends previous work \cite{Vilnis2014} on Gaussian word embeddings to the multimodal joint embedding setting
of images and captions.  From these probabilistic embeddings, we learn a common latent space that captures nuances of meaning with 
respect to image--caption correspondences.  Probabilistic embeddings have also been applied to images alone \cite{Oh2018}.
%
Concurrent work \cite{Chun2021} extends this uni-modal (image-only) model to the same multimodal retrieval setting considered in our paper.
Specifically, \cite{Chun2021} combines ideas from a recent cross-modal architecture \cite{Song2019} with the hedged instance embedding approach of \cite{Oh2018}.
We propose a simpler approach that does not require a specialized loss function to estimate the instance distributions, and show benefits of learning probabilistic embeddings \textit{independently} of the backbone architecture.
As such, we integrate our approach directly with two popular cross-modal architectures and show that both benefit from probabilistic embeddings.
%
Our extensive experimental results point to a clear advantage of our probabilistic
embeddings over point-based embeddings, as well as the method of
\cite{Chun2021}, across all performance metrics. 

%% file: model.tex
\begin{figure}
\centering
	\includegraphics[width=3.5in]{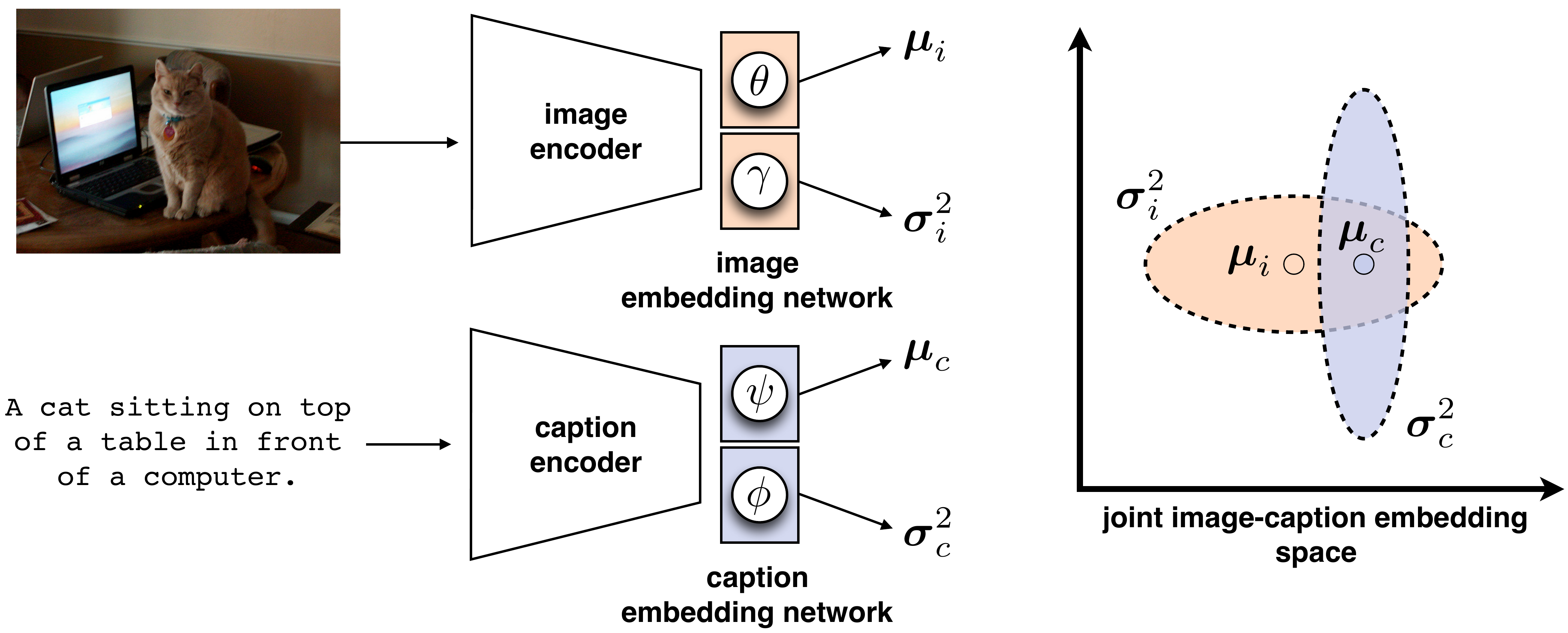}\vspace{5pt}
\caption{Representation overview. Each modality (image/caption) is mapped to a Gaussian distribution
in the joint embedding space shown on the right.}
	\label{fig:architecture}
	\vspace{-15pt}
\end{figure}

In this section, we describe our approach to learning density-based representations of images and captions in a joint embedding space. 
Fig.\ \ref{fig:architecture} provides an overview of our approach.

\subsection{Model}


We use Gaussian distributions, $\mathcal{N}(\vmu,\Sigma)$, with mean $\vmu$ and covariance $\Sigma$, to represent each image and caption in
the joint embedding space. Following \cite{Vilnis2014}, we assume either a diagonal multivariate ellipsoidal or Gaussian distribution to model our probabilistic representations, and learn mappings to the mean and variance vectors.
Let \(f(\cdot)\) and \(g(\cdot)\) denote the initial image and text encodings, 
respectively,
and \(f_\mathrm{emb}(\cdot;\vtheta)\) and \(g_\mathrm{emb}(\cdot;\bm{\psi})\) represent the image and caption embedding functions, respectively, with learnable parameters \(\vtheta\) and \(\bm{\psi}\). We take the output of these functions to learn the means and use the same functions with different learnable parameters \(\bm{\gamma}\) and \(\bm{\phi}\) to learn the variances of the Gaussian representations. Formally,
\begin{align}
\vmu_i & = f_\mathrm{emb}( f(i);\vtheta), &
\vsigma^2_i  &=  f_\mathrm{emb}( f(i);\bm{\gamma}), \\
\vmu_c &=  g_\mathrm{emb}(g(c);\bm{\psi}), &
\vsigma_c^2  &=  g_\mathrm{emb}(g(c );\bm{\phi}), 
\end{align}
where the means and variances \(\vmu_i\), \(\vsigma^2_i\), \(\vmu_c\), \(\vsigma^2_c\) $\in \mathbb{R}^D$, and \(i\) and \(c\) denote the input image and caption, respectively.

With this formulation, our probabilistic approach can be easily adapted to point-based image--text models with minor modifications. Specifically, we can use the point-based embeddings from each modality as our mean outputs, and duplicate the learnable part of the architecture to learn the variances (see Fig.~\ref{fig:architecture}). We can then replace the vector-based similarity metric with a \textit{probabilistic} one (see Section~\ref{ss:trn-loss}) without requiring any changes to the loss function, to learn the image and caption embeddings in the joint space. Please refer to Section~\ref{sec:implementation_details} for a more detailed discussion.
\subsection{Training loss}\label{ss:trn-loss}

Our multimodal representation is trained using pairs of images and corresponding captions, \((i_n, c_n)\) for \(n \in \{1,\dotsc,N\}\), where \(N\) is the total number of ground truth pairs in the dataset. A pair \((i_j,c_k)\) is designated as \textit{matching} if \(j=k\) and \textit{non-matching} if \(j \neq k\). For ease of exposition, we use \((i,c)\) to refer to the embeddings of a matching pair and \((i',c)\) and \((i,c')\) to represent the embeddings of a non-matching pair. Our model is trained using a contrastive loss to learn a joint (image and caption) embedding space in which the elements of a matching pair are closer to each other compared to those of non-matching pairs. 
For the contrastive loss,
we use the hinge-based triplet ranking loss \cite{Chechik2010} with semi-hard negative mining \cite{Schroff2015}: 
\begin{equation}
	\begin{aligned}
		\label{eq:hinge_loss}
		\mathcal{L}(i,c) = \underset{c'}{\mathrm{max}}\left[ \alpha + \mathrm{sim}(i,c') - \mathrm{sim}(i,c) \right]_+ \\
		+\underset{i'}{\mathrm{max}}\left[ \alpha + \mathrm{sim}(i',c) - \mathrm{sim}(i,c) \right]_+,
	\end{aligned}
\end{equation}
where \([x]_+ \triangleq \mathrm{max(x,0)}\), \(\alpha\) is a hyperparameter for the margin,
and $\mathrm{sim}(\cdot,\cdot)$ measures the similarity between two distributions, i.e., the image and caption embeddings.
The total loss is the sum of \(\mathcal{L}(i,c)\) over all positive pairs in the batch, i.e., $\mathcal{L} =  {\sum_{(i,c)}}{\mathcal{L}(i,c)}$.




There are a variety of similarity measures over distributions, $\mathrm{sim}(\cdot,\cdot)$, that can be considered
for our loss, (\ref{eq:hinge_loss}).
In this work, we consider three standard similarity measures, namely, negative Kullback-Liebler (KL) divergence, negative minimum KL divergence, and negative 2-Wasserstein distance; please
see the supplemental for their definitions. 

\subsection{Per-instance measure of uncertainty}

A key advantage of probabilistic representations is that they can express the uncertainty in the input. We propose an explicit measure of uncertainty for an input instance $x$ (a caption or an image),
calculated as the log determinant of the covariance matrix of the learned Gaussian distribution for $x$, i.e., $\mathrm{log}(\mathrm{det}(\Sigma_x))$. 
Based on this definition, the uncertainty of an instance corresponds 
to the overall variance of its learned distribution in the joint space.
We hypothesize that this variance reflects the level of cross-modal ambiguity for an image/caption. That is, the lower the ambiguity of an instance, the lower the uncertainty of its learned representation in the joint space. 
We provide empirical evidence for this hypothesis (see Sec.\ \ref{ss:res-uncertainty}), suggesting that it is indeed through an explicit capturing of the cross-modal ambiguity that probabilistic representations are able to perform more accurate cross-modal retrieval.



%% file: results.tex
\subsection{Datasets}

We report retrieval results on the standard MS-COCO image--caption dataset \cite{Lin2014}.
We use the splits from \cite{Karpathy2015}, containing $82,783$ training, $5000$ validation, and $5000$ test images, where each image is associated with five manually-provided captions. 
Following prior work \cite{Faghri2017,Chun2021,Song2019}, we augment our training data with  
$30,504$ images (and their captions) from the original validation set of MS-COCO that were left out in the split of \cite{Karpathy2015}. We also report retrieval results on the smaller and noisier Flickr30K~\cite{Young2014} dataset. 
Flickr30K contains $31,000$ images each paired with five manually-annotated captions. Following~\cite{Lee2018}, we set aside $1000$ image--caption pairs each for validation and test, and use the rest for training.

Prior work has noted a problem with MS-COCO for cross-modal retrieval evaluation: most pairs are considered as non-matching due to the sparseness of the annotations \cite{Hodosh2016,Hu2019BISON,Parekh2021}.
To address this issue, \cite{Parekh2021} introduced the Crisscrossed (CxC) dataset
%
which extends MS-COCO with additional human annotations for semantic similarities on the validation and test sets.
We use the CxC annotations on the test set as additional evaluation data for the models trained on MS-COCO. 
Following \cite{Parekh2021}, we identify \(10,614\) additional positive pairs based on the annotations, and add them to the original \(25,000\) MS-COCO positive test pairs.

Finally, we evaluate on finer-grained retrieval using Visual Genome \cite{KrishnaVG2016}.
Visual Genome is a large, densely annotated image--caption dataset, 
with images containing tight bounding boxes around objects
and a corresponding short caption describing the bounding box region. The dense annotations 
provide rich information about images, enabling us to investigate the relationship between the complexity of an image/caption and its representation uncertainty. 


\subsection{Implementation details}
\label{sec:implementation_details}

Our models are implemented in PyTorch \cite{PyTorch}.
To show the generality of our probabilistic representation, we adapt two distinct retrieval models: VSE++~\cite{Faghri2017} and VSRN~\cite{Li2019}. Both models learn image--caption embeddings \emph{without} 
cross-domain attention. 
VSE++ is representative of a class of end-to-end models that learn a single embedding from the raw images/captions, and is widely used as a baseline for comparison in cross-modal retrieval. 
VSRN extracts a set of image features from regions identified by an object detector, and is 
among the best performing retrieval models that does not involve cross-domain attention. For both models, we use the official codebase.


As mentioned earlier, a key advantage of our probabilistic approach is that it can be adapted to any image--text model with simple modification to the original model. Specifically, we take the output of the original model as the mean. We learn the variance by duplicating the learnable part of the model and thus generate another output of the same dimension as the variance vector. Next, we describe our
modifications to VSE++ and VSRN.

\noindent\textbf{Image representations. } 
For VSE++, to compute the mean, 
we linearly project the image features, $f(i) \in \R^{2048}$, generated by ResNet-152~\cite{resnet152}, yielding $\mathbf{\mu}_{i} \in \R^{1024}$. For VSRN, we use a recurrent network to consolidate the contextualized regions features, $f(i)\in \R^{36\times 2048}$, generated by bottom-up attention \cite{Anderson2018CVPR} followed with a graph neural network.  We take the final hidden state of the recurrent network as the mean vector, $\mathbf{\mu}_{i} \in \R^{2048}$. As mentioned earlier, we duplicate the learnable embedding layers of the mean network with no parameter sharing to generate the variance vectors, yielding $\mathbf{\sigma}_{i} \in \R^{1024}$ for VSE++ and $\mathbf{\sigma}_{i} \in \R^{2048}$ for VSRN.
Unless otherwise stated, we do not 
fine-tune the image backbones.


\noindent\textbf{Caption representations. } For our probabilistic versions of both VSE++ and VSRN, we use randomly initialized 300-dimensional embeddings to represent each word in a caption. Word embeddings are then passed through a unidirectional Gated Recurrent Unit (GRU) \cite{gru}. The final hidden state of the GRU is taken as the caption mean embedding in the joint space, with $\mathbf{\mu}_{c} \in \R^{1024}$ for VSE++ and  $\mathbf{\mu}_{c} \in \R^{2048}$
for VSRN. For the variances, we duplicate the GRU layer with no parameter sharing in both VSE++ and VSRN.

 
 \noindent\textbf{Training details. } 
 We train all our models for $30$ epochs with a learning rate of \(2\mathrm{e}{-4}\) and a learning rate decay factor of $10$ at epoch $15$. We use the Adam optimizer~\cite{Kingma2015} with a mini-batch size of $128$. For the loss function, (\ref{eq:hinge_loss}), we use a margin of \(0.2\) for VSE++ and \(0.1\) for VSRN. 
 To ensure numerical stability, we estimate the log variances instead of variances, and following previous work \cite{Vilnis2014} we bound all variance values to the range \([0.1,10]\). For modeling our uncertainties, we consider both 
 ellipsoidal and spherical Gaussian distributions. We estimate the variances of the spherical distributions in two ways: (i) Learn an ellipsoidal distribution and take the average of the variances as the constant spherical variance (spherical-avgpool); or ii) learn a single constant (spherical-one value). 
%
Since our probabilistic models include additional parameters, 
we also 
report results for the original models with increased capacity (doubling the dimension of their representations in the joint space) to roughly match the capacity of our models.


\subsection{Evaluation measures}

Following prior work on cross-modal retrieval, 
we evaluate our models on image-to-text and text-to-image 
tasks, and report R(ecall)@$K$ (with $K = 1, 5, 10$). R@$K$ is 
the percentage of queries for which the top-$K$ retrieved samples contains the correct (matching) item. We choose the best models based on the sum of the recall values (rsum) on the validation sets. 

Recall measures do not penalize or reward a model based on how good of a match the rest of the top-$K$ retrieved samples are. As explored in \cite{Parekh2021} with CxC, there are many more plausible matches in a dataset like MS-COCO than the annotated ones. Thus, previous work has proposed R-Precision to complement R@$K$ values~\cite{Musgrave2020,Chun2021}, calculated as the ratio of all positive items in the top-$r$ retrieved samples for a given query (where $r$ is the number of ground truth matches), averaged over test queries.

We report two versions of the R-Precision, each relying on a different source to determine ground truth matches that are not available in the original image--caption datasets. We report Plausible Match R-Precision (PMRP) on MS-COCO, where we determine the plausible ground truth matches based on the overlap of human-labeled object classes on the MS-COCO images, as in~\cite{Chun2021}. More concretely, each image and its associated captions are assigned a binary label vector indicating the appearance of a set of object classes in the image. We consider an image--caption pair as a plausible match if their corresponding binary label vector differs at most by a given number of positions, $\zeta$.
Following ~\cite{Chun2021}, we evaluate on $\zeta \in \{0,1,2\}$ and take their average. The drawback of PMRP is that it heavily relies on exhaustive and accurate object annotations, which is difficult to collect. In addition, the assumption that plausible matches can be determined based on the overlap in objects may not always be valid. 
Specifically, cases can arise where the presence of the same object classes in two images can have different meanings.
We thus propose an alternative way to calculating R-Precision, termed 
\(\mathrm{RPC^2}\), that relies on the additionally-collected human annotations in CxC to identify ground-truth plausible matches.
Since measuring precision requires special annotations, we can only provide them on MS-COCO, and its extension, CxC.
Following previous evaluations on Flickr30k, we only report recall values.


\subsection{Ablation study}

 In the supplemental, we provide an extensive ablation study on the MS-COCO and Flickr30K validation sets over the similarity metric used in our loss, (\ref{eq:hinge_loss}), and the form of the 
covariance (i.e., spherical vs.\ ellipsoidal). 
Results reported in the remainder of this paper are based on the best combination for each model and dataset, as explained in the next paragraph.

On the MS-COCO dataset, all models perform best with ellipsoidal covariance matrices.
We get the best retrieval performance for VSE++ (no fine-tuning) and VSRN
with the Wasserstein metric (with 4\% difference in rsum between the best-performing and the worst-performing model for VSE++, and 1\% difference for VSRN). For the fine-tuned VSE++ (VSE++ft), using the minimum KL metric yields the highest performance (with a 5\% difference in rsum). 
On Flickr30K, the best results for all models are obtained with spherical covariance matrices.
The best performance is achieved by using KL divergence for VSE++ (a 2\% difference in rsum) and Wasserstein for VSRN (a 3\% difference in rsum).



\begin{table*}[t]
	\begin{center}
		\scriptsize{
			\begin{tabular}{@{}lllllllll|llllllll@{}}
				\toprule
				\multirow{3}{*}{Model} & \multicolumn{8}{c|}{1K (averaged over five fold)} & \multicolumn{8}{c}{5K (full test set)}\\
				 & \multicolumn{4}{c}{image-to-text} & \multicolumn{4}{c|}{text-to-image}& \multicolumn{4}{c}{image-to-text} & \multicolumn{4}{c}{text-to-image} \\
				
				& PMRP & R1 & R5 & R10 & PMRP &  R1 & R5 & R10& PMRP & R1 & R5 & R10 & PMRP &  R1 & R5 & R10 \\
				\hline
				VSE++ \(^*\) & 39.2 & 58.3 &	86.1	&93.3	& 39.3&43.7&	77.6&	87.8& 29.3 & \textbf{35.3} & 63.3 & 75.3 & 29.1 & 23.2 & 50.3 & 63.5 \\
				VSE++ \(^\dagger\) & 39.2 & 59.0 &	\textbf{87.0} &	\textbf{94.2}& 33.3 & 44.7	& 78.4	& 88.3& 29.1 & 33.8&\textbf{64.6}&76.1& 28.9 &23.3&51.2&64.1 \\
				VSE++ ours & \textbf{41.1}& \textbf{59.3}	& {86.8}	& {93.8}	& \textbf{41.7} &	\textbf{45.6 }&	\textbf{79.2} &	\textbf{89.1}& \textbf{31.3}& {34.7} & \textbf{64.6} &	\textbf{76.6} & \textbf{31.2} & \textbf{24.5} & \textbf{52.4} & \textbf{65.6}\\
				\hline
				VSE++ft~\cite{Faghri2017} & 40.6 & 64.6 & 90.0 & 95.7 & 41.2 & 52.0 & 84.3 & 92.0 & 29.8 & 41.3 & 71.1 & 81.2 & 30.0 & \textbf{30.3} & 59.4 & 72.4 \\
				VSE++ft ours & \textbf{43.0} & \textbf{66.4} & \textbf{91.2} & \textbf{96.5} & \textbf{43.8} & \textbf{52.5} & \textbf{84.6} & \textbf{92.7} & \textbf{32.3} & \textbf{41.4} & \textbf{72.4} & \textbf{82.9} & \textbf{32.3} & 30.0 & \textbf{60.4} & \textbf{72.7} \\
				\hline
				VSRN  \(^*\)& 41.2 & \textbf{74.0}	& {94.3}	& \textbf{97.9} & 42.4 & 60.8&	88.4&	94.1	& 29.7 & 50.2 & 79.6 & 87.9 & 29.9 & 37.9 & 68.5 & 79.4  \\
				VSRN \(^\dagger\) &39.5 & 70.5 &	93.1 &	97.1 & 40.2 & 58.8	&88.2	& 94.5& 27.7 & 45.3 & 76.0 & 86.4 &28.1  & 35.9 &66.7 & 78.3 \\
				VSRN ours & \textbf{45.8} & 	{73.8} &	\textbf{94.4}	& \textbf{97.9} & \textbf{46.7} & \textbf{61.3} &	\textbf{89.2} &	\textbf{95.2}& \textbf{34.2} & \textbf{51.1} &\textbf{80.1} & \textbf{89.4} & \textbf{34.5} & \textbf{38.8} & \textbf{69.1} & \textbf{80.2} \\
				\hline
				\hline
				PCME~\cite{Chun2021} & 45.1 & 68.8 & 91.6 & 96.7 & 46.0 & 54.6 & 86.3 & 93.8& 34.1 & 44.2 & 73.8 & 83.6 & 34.4 & 31.9 & 62.1 & 74.5 \\
				PVSE~\cite{Song2019} &42.8 & 69.2 & 91.6 & 96.6 & 43.6 & 55.2 & 86.5 & 93.7 & 31.8 & 45.2 & 74.3 & 84.5 & 32.0 &32.4 & 63.0 & 75.0\\
				\bottomrule
			\end{tabular}
		}
	\end{center}
	\vspace{-10pt}
	\caption{Retrieval results on MS-COCO.  
	\(^*\) denotes results generated with the official saved models; \(^\dagger\) identifies the higher capacity models trained using the official code, and ft stands for fine-tuned. Best results in each group are highlighted in \textbf{bold}.}
	\vspace{-10pt}
	\label{tab:results_coco}
\end{table*}

\begin{table*}[t]
	\begin{center}
		\scriptsize{
			\begin{tabular}{@{}lllllllll|llllllll@{}}
								\toprule
				\multirow{3}{*}{Model} & \multicolumn{8}{c|}{1K (averaged over five fold)} & \multicolumn{8}{c}{5K (full test set)}\\
				 & \multicolumn{4}{c}{image-to-text} & \multicolumn{4}{c|}{text-to-image}& \multicolumn{4}{c}{image-to-text} & \multicolumn{4}{c}{text-to-image} \\
				
				& \(\mathrm{RPC^2}\) & R1 & R5 & R10 & \(\mathrm{RPC^2}\) &  R1 & R5 & R10& \(\mathrm{RPC^2}\) & R1 & R5 & R10 & \(\mathrm{RPC^2}\) &  R1 & R5 & R10 \\
				\hline
			VSE++ \(^*\) & 39.0 & 59.3 & 87.1 &	93.9&	40.2& 44.7 &	78.4	& 88.4 & 23.8 & \textbf{37.2}	& 66.7	& 78.6	&23.4& 25.3	& 54.0 & 	67.3\\
			VSE++ ours  & \textbf{39.3} & \textbf{60.3}	& \textbf{87.6}	& \textbf{94.2} & \textbf{42.1}	& \textbf{46.6} &	\textbf{79.9}	& \textbf{89.5 }& \textbf{24.1}  & {36.7} & \textbf{68.0}	& \textbf{80.0}	& \textbf{24.8}& \textbf{26.5}	&\textbf{ 55.8}	& \textbf{69.0}\\
			\hline
			VSE++ft \(^*\) & 44.1 & 65.6 & 90.9 & 96.1 &47.8 & 53.2 & 84.9 & 92.3 &  \textbf{28.5} & 43.3 & 74.2 & 84.1 & \textbf{30.1} & \textbf{32.5} & 62.7 &  75.3 \\
			VSE++ft ours \(^*\) & \textbf{44.2} & \textbf{67.6} & \textbf{91.9} & \textbf{96.8} & \textbf{48.0} & \textbf{53.6} & \textbf{85.2} & \textbf{93.1} & 28.1 & \textbf{43.7} & \textbf{74.7} & \textbf{85.0} & 29.9 & 32.1 & \textbf{63.4} & \textbf{75.5}\\
			\hline
			VSRN \(^*\) & {50.4} & \textbf{74.8} &  {94.8} & {98.1} & 54.8 & 61.8 & 88.8 & 94.4& 34.3 &  52.4 & 81.8 & 90.0 & 37.1 & 40.1 &71.1& 81.5\\
			VSRN ours & \textbf{50.5} &	{74.5} 	& \textbf{94.9} &	\textbf{98.2} &	\textbf{55.2} & \textbf{62.3} &	\textbf{89.7} &	\textbf{95.4} & \textbf{34.5} &	\textbf{53.1} &	\textbf{82.6} &	\textbf{91.1} &	\textbf{37.7} & \textbf{40.9} &	\textbf{71.5} &	\textbf{82.4}\\
			\hline
			\hline
				PCME \(^*\)&  45.9 & 69.2 & 92.0 & 97.0 & 49.2 & 55.3 & 86.5 & 94.1 & 29.5 & 45.2 & 75.1 & 85.1 & 30.7 & 33.4 & 64.1 & 76.3\\
				PVSE \(^*\) & 46.4 & 70.0 & 92.2 & 97.2 & 50.3 & 56.4 & 87.0 & 94.0 & 30.3 & 47.1 & 77.2 & 87.0 & 32.1 & 34.6 & 66.0 & 77.8\\
				\bottomrule
			\end{tabular}
		}
	\end{center}
	\vspace{-10pt}
	\caption{Retrieval results on the MS-COCO test set with the additional CxC annotations. 
	\(^*\) denotes results generated using the official saved models and ft stands for fine-tuned. Best results in each group are highlighted in \textbf{bold}.}
	\label{tab:results_cxc}
	\vspace{-10pt}
\end{table*}

\subsection{Retrieval results}\label{ss:res-retrieval}


Table~\ref{tab:results_coco} summarizes the R@$K$ and PMRP results on MS-COCO test pairs. 
Apart from the results on the original models with and without our proposed probabilistic embeddings, we also include published results for PCME~\cite{Chun2021} and PVSE~\cite{Song2019} for direct comparison.
Note, both models propose new architectures and optimize their architecture for cross-modal retrieval. 
In contrast, our approach 
can be easily adapted to any architecture with minimal change.
Moreover, PCME is the only model that
learns  probabilistic  representations  but  it  does  it  at  the  cost  of  retrieval performance.
For PVSE, we report the published recall values in~\cite{Song2019} and the published PMRP values in~\cite{Chun2021}.
As can be seen in Table~\ref{tab:results_coco}, 
adding probabilistic representations (ours) 
to both VSE++ and VSRN yields
improved (or competitive) performance across the board.
As noted in prior work \cite{Musgrave2020,Chun2021,Parekh2021}, the R-Precision measures are more suited at capturing fine-grained differences in the performance of cross-modal retrieval models. PCME improved on VSRN (state of the art on cross-modal retrieval at the time) when comparing precision (PMRP) but not recall. We can see that adding probabilistic representations to VSRN (VSRN ours) results in further improvements on precision over PCME for both image-to-text and text-to-image retrieval, while substantially outperforming PCME in terms of recall.



As previously mentioned, a more accurate measure of precision is the \(\mathrm{RPC^2}\) metric that uses actual human annotations to determine the plausible matches. Table \ref{tab:results_cxc} summarizes our finer-grained retrieval results on  the MS-COCO test set with the additional CxC annotations.
These results are in line with those in Table~\ref{tab:results_coco}, and further show the superiority of our probabilistic representations. Interestingly, although PCME showed improved precision over all non-probabilistic models (VSE++ with and without finetuning, VSRN, and PVSE) with the PMRP metric, it is only better than VSE++ models based on the more accurate precision measure of \(\mathrm{RPC^2}\). Our probabilistic approach added to VSRN achieves the best performance in terms of both precision \textit{and} recall.
\begin{table}
	\begin{center}
        \scriptsize{
        \begin{tabular}{@{}lllllll@{}}
			\toprule
			\multirow{2}{*}{Model} & \multicolumn{3}{c}{image-to-text} & \multicolumn{3}{c}{text-to-image}\\
			& R@1 & R@5 & R@10 &  R@1 & R@5 & R@10\\
			\hline
			VSE++~\cite{Faghri2017} & 	43.7	& 71.9	& 82.1 &32.3 &	60.9 &	72.1 \\
			VSE++ \(^\dagger\)& 45.6 & 73.0 & 82.1 & \textbf{34.1} &\textbf{62.4} & \textbf{73.6}\\
			VSE++ ours & \textbf{48.8} & \textbf{74.5} &	\textbf{83.1} & 33.3	&\textbf{62.4}	& 72.9 \\
			\hline
			VSE++ft~\cite{Faghri2017} & 52.9 & 80.5 & 87.2 & 39.6 & 70.1 & 79.5 \\
			VSE++ft ours & \textbf{56.9} & \textbf{82.5} & \textbf{90.1} & \textbf{41.1} & \textbf{71.9} & \textbf{80.3} \\
			\hline
			VSRN \(^*\) &  \textbf{70.4} &	89.2 &	93.7	&\textbf{53.0}	& 77.9	& 85.7	\\
			VSRN \(^\dagger\) & 	69.1	& \textbf{90.3}	& \textbf{94.4} & 52.3	& 78.5 &	{86.1} \\
			VSRN ours &  69.2	& 89.4 &	{94.1} &52.3 &	\textbf{79.3}	& \textbf{86.5}	\\
			\bottomrule
		\end{tabular}
        }
	\end{center}
	\caption{Flickr30k test set retrieval results.
	\(^*\) denotes results generated with the official saved models and \(^\dagger\) the higher capacity models trained using the official code. Best results in each group are highlighted in \textbf{bold}.}
	\label{tab:results_f30k}
	\vspace{-10pt}
\end{table}

Table~\ref{tab:results_f30k} shows our results on Flickr30K \cite{Young2014}.
Note that the R-Precision measures require special annotations that are available for MS-COCO but not for Flickr30K, and as such cannot be reported here.
We can see that for VSE++, the probabilistic model (ours) performs better than the base model, both with and without fine-tuning. Compared to the high-capacity model, VSE++ ours performs better for image-to-text and is competitive for text-to-image. For VSRN, we do not see a clear advantage for any of the models. Nonetheless, we stress that our model is competitive and has an associated uncertainty measure that can be used a diagnostic tool.

\subsection{Uncertainty and cross-modal ambiguity}\label{ss:res-uncertainty}





\begin{figure}[t]
\captionsetup[subfigure]{justification=centering}
    \centering
    \begin{subfigure}[b]{0.45\textwidth}
        \includegraphics[width=\textwidth]{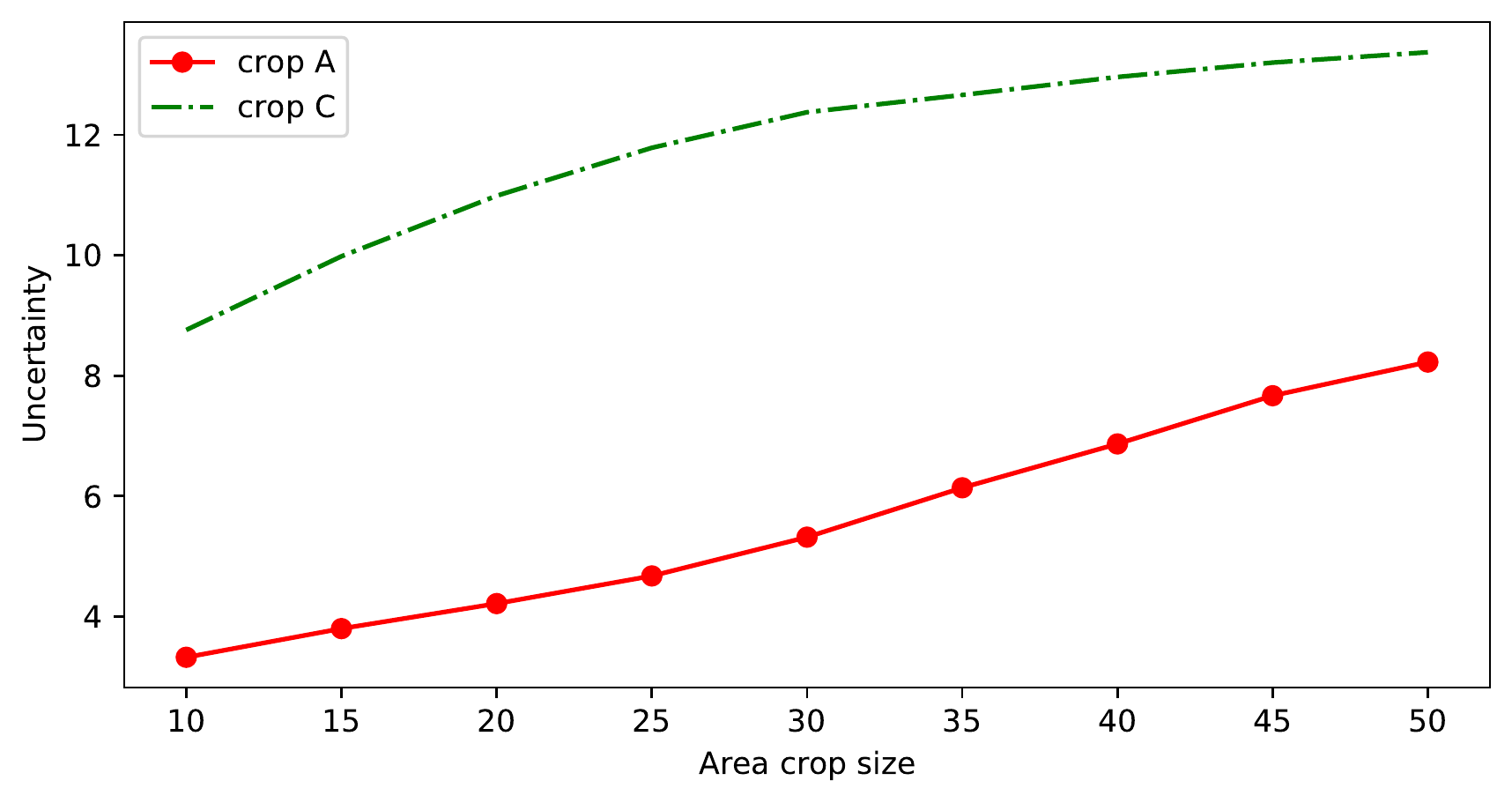}
        \caption{Images}
        \label{fig:2a} 
    \end{subfigure}

    \begin{subfigure}[b]{0.45\textwidth}
        \includegraphics[width=\textwidth]{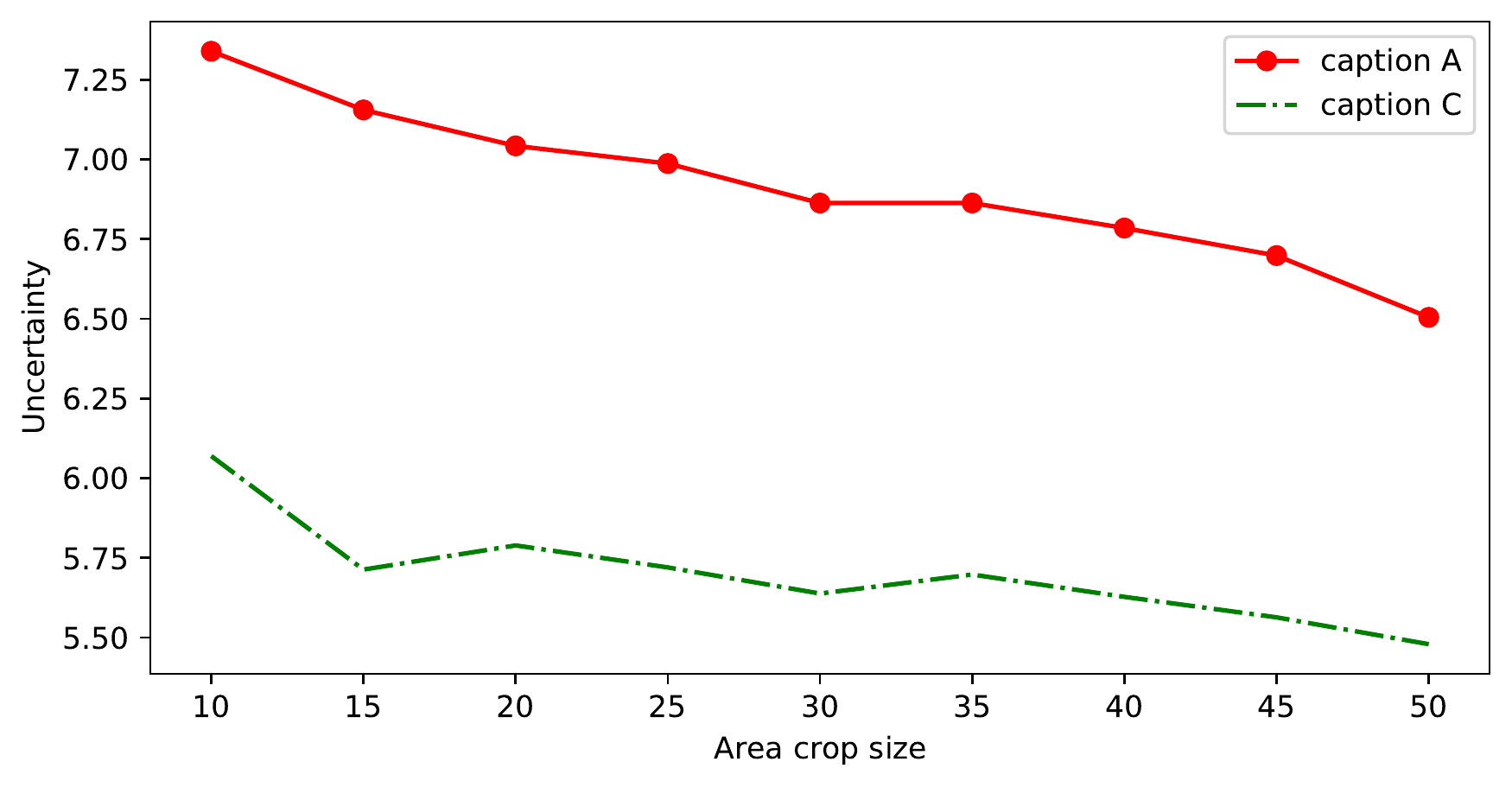}
        \caption{Captions}
        \label{fig:2b}
    \end{subfigure}
\vspace{-5pt}
	\caption{Average uncertainty values of images and captions with varying area thresholds. 
	}
	\label{fig:vg}
	\vspace{-10pt}
\end{figure}

Our guiding hypothesis is that
the uncertainty encoded in our probabilistic embeddings captures the cross-modal ambiguity in the input. For example, an image with several objects/events can be described in many different ways, each description focusing on a subset of the visual scene. 
That is, a content-rich image has a high degree of cross-modal ambiguity. In contrast, an image with a single common object/event has a low degree of cross-modal ambiguity. To verify this hypothesis, we create a controlled dataset using the densely-captioned Visual Genome to show that the uncertainty of an image/caption is indicative of its cross-modal ambiguity. We then investigate the connection between ambiguity and cross-modal retrieval through a controlled binary selection experiment. The rest of this section explains our controlled dataset and experiments in more detail.

\begin{figure*}
    \begin{center}
        \includegraphics[width=\textwidth]{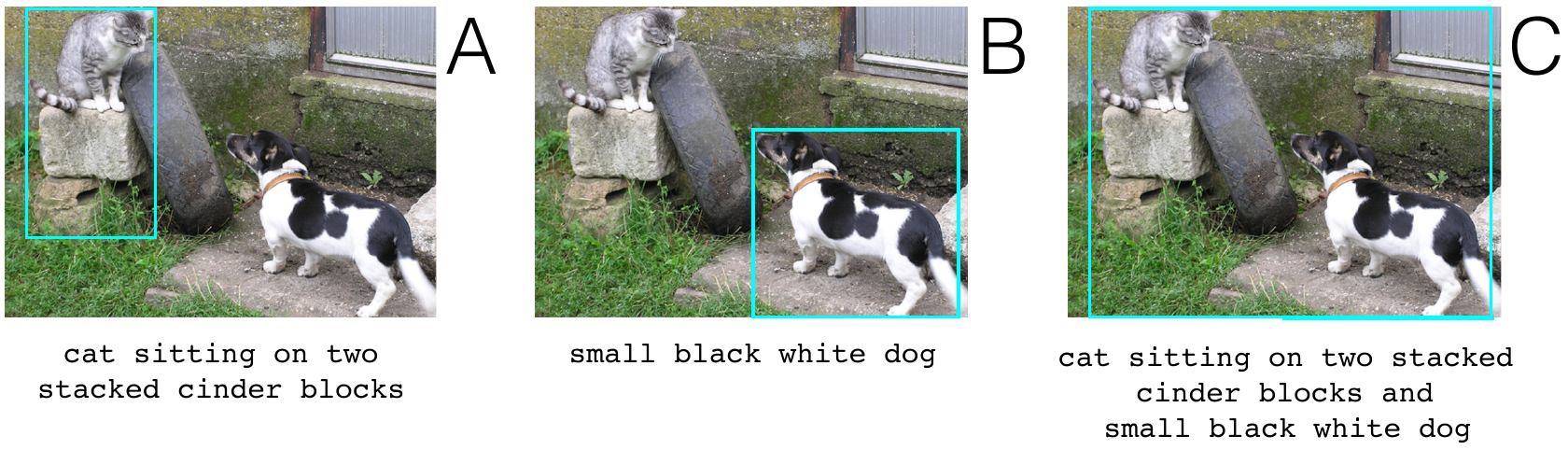}
        \vspace{1pt}
    	\captionof{figure}{An example of our dataset preparation for the uncertainty experiments.
    	A and B denote the first and second crop, respectively, and C their union. }
    	\label{fig:vg_example2}
    \end{center}
\end{figure*}

Our goal is to compare uncertainties for pairs of images (captions) where one is clearly less ambiguous than the other. 
To form such pairs, we randomly sample $2000$ images from Visual Genome. From each image, we generate a triplet of crops (A, B, C) as follows: For a given area threshold, we select the largest ten crops whose area is less than the threshold. Each of these crops is considered to be fairly unambiguous, since it was created by assigning a tight bounding box around a detected object~\cite{KrishnaVG2016}. We assign the largest among these as the first item in the triplet (crop A). From the remaining nine crops, we choose crop B as the one with the smallest overlap (measured as IoU) with crop A. Crop C is formed by taking the union of crops A and B. 
Note that this process results in a triplet of differently-sized crops of the same image, formed in a way that ensures crop C contains more objects (and as such is more ambiguous) than crop A. We choose triplets from the same image to control for other visual properties of an image that may contribute to its ambiguity. For corresponding captions we use the Visual Genome annotation for crop A and B. To create a caption for crop C, we concatenate the captions of crops A and B (whose union forms crop C) with the word ``and''. Fig.~\ref{fig:vg_example2} shows an example of our cropping process.

To further understand how the connection between cross-modal ambiguity and uncertainty affects cross-modal retrieval, we perform a controlled experiment over the crop triplet dataset that we created as explained above.  Following prior work suggesting multiple-choice selection as a finer-grained alternative to retrieval for the evaluation of image--text matching models \cite{Hodosh2016,Shekhar2017,Hu2019BISON}, we frame our experiment as a binary selection task.
Specifically, we present each model (either the original VSE++ or VSE++ with probabilistic embeddings) with either caption (crop) A or C as the query, and select between crop (caption) A and C based on their similarities with the query in the joint space. 

\begin{table}
    \begin{center}
    \scriptsize{
    \begin{tabular}{@{}l|ll!{\vrule width 1pt}ll@{}}
    \toprule
         \multirow{2}{*}{model} & \multicolumn{2}{c!{\vrule width 1pt}}{image-to-text} &\multicolumn{2}{c}{text-to-image}   \\
          & crop A & crop C & caption A & caption C \\
         \hline
         VSE++* & \textbf{59.7} & 66.0 & \textbf{74.8} & 54.0 \\
         VSE++ ours & 58.6 & \textbf{69.5} & 71.7 & \textbf{58.9} \\
    \bottomrule
    \end{tabular}
    }
    \end{center}
    \caption{Binary selection accuracy over the $2000$ Visual Genome crop triplets.}
    \label{tab:vg_detail2}
\end{table}

Fig.~\ref{fig:vg}(a) plots the average uncertainty of crops A and C for different area thresholds ranging from $10\%$ to $50\%$, measured using our probabilistic VSE++ model.
As can be seen, irrespective of the area threshold, the uncertainty of the union crop (which is expected to contain more objects and be more ambiguous) is always notably higher, as expected. This corresponds to image-to-text binary selection accuracy in Table~\ref{tab:vg_detail2}: Looking at the image-to-text columns, we can see that for an unambiguous crop such as A used as the query, the models perform comparably. In contrast, for a more complex and ambiguous crop such as C, our probabilistic model substantially improves upon its point-based embedding counterpart, suggesting that explicitly capturing uncertainty is beneficial.

Fig.~\ref{fig:vg}(b) plots the average uncertainty of captions A and C for the same range of area threshold values as above. Here, we see a different pattern: the uncertainty of the conjoined caption (caption C corresponding to the crop with more objects) is lower than that of caption A. 
This is expected because the conjoined caption provides more information about the image it can describe, and as such is expected to have a lower degree of cross-modal ambiguity. This corresponds to the text-to-image binary selection accuracy in Table~\ref{tab:vg_detail2}: As can be seen in the the text-to-image columns, we do not see an advantage for the probabilistic representations when the task is to select the correct crop given the short caption A as the query. In contrast, for the conjoined caption C, 
our probabilistic model shows an advantage in selecting the correct crop.

%% file: conclusion.tex
In this paper, we proposed a simple yet general and effective approach for learning probabilistic representations for cross-modal retrieval models.  We demonstrated the generality of our method by extending extant and distinct
point-based cross-modal retrieval models.
Through extensive experimentation, we showed that our probabilistic representations yield superior or competitive performance to their point-based analogs
and is superior to the lone, concurrent work proposing probabilistic embeddings \cite{Chun2021}.
Further, our embeddings have 
the added benefit over point-based ones by providing a measure of uncertainty that  we empirically validated
to capture the cross-modal ambiguity in images and captions.

%% file: supplemental.tex
\subsection{Similarity metrics}
Below, we provide details on the similarity metrics that we use in this work:
\begin{enumerate}
	\item \textbf{Negative Kullback-Liebler (KL) divergence}. This metric is the only asymmetric similarity metric that we use. Since images contain more details than captions (i.e., a caption cannot verbalize all aspects in an image), we expect the captions to have higher uncertainties.  Therefore, we take the KL divergence with respect to the caption distribution, i.e., \( \mathrm{KL}(\gN(\vmu_i,\Sigma_i) ||\gN(\vmu_c,\Sigma_c))\), with the following closed form:
	\begin{equation}
		\label{eq:kl}
		\begin{aligned}
			-\frac{1}{2}\biggl(\mathrm{tr}(\Sigma_c^{-1}\Sigma_i) {} + (\vmu_i-\vmu_c)^\mathrm{T}\Sigma_c^{-1}(\vmu_i-\vmu_c) - D  \\ + \mathrm{ln}{\left(\frac{\mathrm{det}(\Sigma_c)}{\mathrm{det}(\Sigma_i)}\right)} \biggr),
		\end{aligned}
	\end{equation}
	where $\mathrm{tr}(\cdot)$ and $\mathrm{det}(\cdot)$ are the trace and determinant matrix operators, respectively. Since we assume diagonal covariance matrices, (\ref{eq:kl}) can be simplified to 
    \begin{align}
	    -\frac{1}{2}\mathrm{tr}\left(\mathrm{diag}\left(\frac{\vsigma_i^2}{\vsigma_c^2} 
	    - \mathrm{ln}\frac{\vsigma_i^2}{\vsigma_c^2}
	    + \frac{(\vmu_i-\vmu_c)^2}{\vsigma_c^2} -\bm{1}\right)\right),
	\end{align}
    where all vector operations are element-wise, \(\bm{1}\) is a $D$-dimensional vector of all ones and \(\mathrm{diag}(\va)\) is a diagonal matrix, with \(\va\) containing the elements on its diagonal.  For completeness, we also considered taking the KL divergence with respect to the image embeddings and
    observed a performance drop in retrieval, as expected.  Please see Section~\ref{sec:supp_ablations} in this supplemental for more details.
    


	\item \textbf{Negative Minimum KL divergence.} We define this measure as a symmetric variant of the KL divergence:
	\begin{equation}
		\begin{aligned}
		\label{eq:minkl}
			-\mathrm{min}\{\mathrm{KL}(\gN(\vmu_i,\Sigma_i) ||\gN(\vmu_c,\Sigma_c)), \\ \mathrm{KL}(\gN(\vmu_c,\Sigma_c)||\gN(\vmu_i,\Sigma_i))\}.
		\end{aligned}
	\end{equation}
	%
	\item \textbf{Negative 2-Wasserstein distance}. The Wasserstein distance is a symmetric distance metric between probability distributions. We use the second-order Wasserstein (2-Wasserstein) because it admits the following
	closed-form for Gaussians \cite{Mallasto2017}:
	\begin{align}
	\label{eq:wasserstein}
		-\sqrt{\|\vmu_i-\vmu_c\|^2 + \mathrm{tr}{\left(\Sigma_i+\Sigma_c-2\left(\Sigma_i^{\frac{1}{2}}\Sigma_c\Sigma_i^{\frac{1}{2}}\right)^{\frac{1}{2}}\right)}},
	\end{align}
	where \(\|\cdot\|\) is the Euclidean norm operator. 
	Given our assumption of diagonal covariance matrices, (\ref{eq:wasserstein}) simplifies to
    \begin{align}
        -\sqrt{\|\vmu_i-\vmu_c\|^2 
        + \|\vsigma_i^2-\vsigma_c^2\|^2}.
    \end{align}
\end{enumerate}

\subsection{Ablations}
\label{sec:supp_ablations}
We provide ablations on the similarity metrics and distribution shapes on the MS-COCO validation set in
Table~\ref{tab:new_spherical_dev}. 
We estimate the variances of the spherical distributions in two ways: (i) Learn an ellipsoidal distribution and take the average of the variances as the constant spherical variance (spherical-avgpool); or ii) learn a single constant (spherical-one value).
As we can see, all models perform best with ellipsoidal distributions. For VSE++ (no fine-tuning) and VSRN, the Wasserstein metric ((8) in the main manuscript) shows the best performance in terms of sum of recalls (rsum), while for VSE++ft the best results are achieved with minimum KL.

\begin{table*}
	\begin{center}
        \scriptsize{
        \begin{tabular}{@{}llllllllll@{}}
			\toprule
			\multirow{2}{*}{model} &\multirow{2}{*}{metric} &\multirow{2}{*}{shape}& \multicolumn{3}{c}{image-to-text} & \multicolumn{3}{c}{text-to-image} & \multirow{2}{*}{rsum}\\
			& & & R@1 & R@5 & R@10 &  R@1 & R@5 & R@10\\
			\midrule
			VSE++ ours & KL~(\ref{eq:kl}) & spherical-avgpool & 57.2 & 86.5 & 94.0 & 43.9 & 78.1 & 88.9 & 448.6 \\
			VSE++ ours & KL~(\ref{eq:kl}) & spherical-one value & 56.2 & 86.5 & 94.1 & 43.5 & 77.7 & 88.8 & 446.8\\
			VSE++ ours & KL~(\ref{eq:kl}) & ellipsoidal & 61.3 & 88.1 & 94.1 & 46.6 & 80.6 & 90.4 & 461.1\\
			VSE++ ours & minKL~(\ref{eq:minkl})& spherical-avgpool & 54.7 & 87.2 & 94.7 & 43.7 & 77.9 & 88.7 & 446.9\\
			VSE++ ours & minKL~(\ref{eq:minkl})& spherical-one value & 57.1 & 87.6 & 93.8 & 43.8 & 77.7 & 88.8 & 448.8\\
			VSE++ ours & minKL~(\ref{eq:minkl})& ellipsoidal & 60.0 & 87.6 & 95.2 & 46.8 & 80.9 & 91.0 & 446.9\\
			VSE++ ours & Wasserstein~(\ref{eq:wasserstein}) & spherical-avgpool & 58.1 & 88.3 & 94.3 & 44.9 & 79.8 & 89.3 & 454.7\\
			VSE++ ours & Wasserstein~(\ref{eq:wasserstein}) & spherical-one value & 59.3 & 87.5 & 94.4 & 45.0 & 79.4 & 90.1 & 455.7\\
			VSE++ ours & \textbf{Wasserstein~(\ref{eq:wasserstein})} & ellipsoidal & 61.3 & 88.3 & 95.2 & 47.1 & 81.0 & 91.0 & 463.9\\
			\midrule
			VSE++ft ours & KL~(\ref{eq:kl}) & spherical-avgpool & 68.8 & 92.4 & 97.3 & 53.7 & 86.4 & 93.7 & 492.3\\
			VSE++ft ours & KL~(\ref{eq:kl}) & spherical-one value & 61.7 & 90.1 & 95.7 & 49.3 & 82.9 & 92.3 & 472.0\\
			VSE++ft ours & KL~(\ref{eq:kl}) & ellipsoidal & 70.0 & 92.8 & 96.3 & 54.3 & 86.8 & 93.7 & 493.9\\
			VSE++ft ours & minKL~(\ref{eq:minkl})& spherical-avgpool & 62.4 & 90.5 & 96.1 & 50.0 & 84.4 & 92.6 & 476.0\\
			VSE++ft ours & minKL~(\ref{eq:minkl})& spherical-one value & 61.4 & 90.1 & 95.6 & 49.1 & 83.1 & 91.9 & 471.2\\
			VSE++ft ours & \textbf{minKL~(\ref{eq:minkl})}& ellipsoidal & 68.8 & 93.2 & 97.6 & 55.4 & 86.9 & 93.5 & 495.4\\
			VSE++ft ours & Wasserstein~(\ref{eq:wasserstein}) & spherical-avgpool& 61.8 & 90.3 & 95.5 & 52.8 & 85.2 & 93.1 & 478.7\\
			VSE++ft ours & Wasserstein~(\ref{eq:wasserstein}) & spherical-one value & 67.9 & 92.6 & 97.1 & 54.2 & 86.3 & 93.8 & 491.9 \\
			VSE++ft ours & Wasserstein~(\ref{eq:wasserstein}) & ellipsoidal& 67.3 & 93.1 & 97.4 & 54.2 & 86.7 & 93.8 & 492.5\\
			\midrule
			VSRN ours & KL~(\ref{eq:kl}) & spherical-avgpool & 76.5 & 95.9 & 98.5 & 62.2 & 90.0 & 96.0 & 519.1\\
			VSRN ours & KL~(\ref{eq:kl}) & spherical-one value & 77.5 & 96.7 & 98.7 & 62.0 & 89.9 & 95.7 & 520.5\\
			VSRN ours & KL~(\ref{eq:kl}) & ellipsoidal & 76.5 & 96.3 & 98.1 & 62.9 & 91.0 & 95.9 & 520.7\\
			VSRN ours & minKL~(\ref{eq:minkl})& spherical-avgpool & 76.0 & 96.0 & 98.3 & 63.5 & 90.9 & 96.4 & 521.1\\
			VSRN ours & minKL~(\ref{eq:minkl})& spherical-one value & 75.3 & 96.2 & 98.5 & 60.8 & 90.7 & 96.2 & 517.7\\
			VSRN ours & minKL~(\ref{eq:minkl})& ellipsoidal & 77.1 & 96.2 & 98.8 & 63.6 & 90.5 & 96.2 & 522.4\\
			VSRN ours & Wasserstein~(\ref{eq:wasserstein}) & spherical-avgpool	& 76.8 & 96.8 & 99.2 & 63.2 & 90.5 & 96.0 & 522.5\\
			VSRN ours & Wasserstein~(\ref{eq:wasserstein}) & spherical-one value & 76.4 & 96.5 & 98.9 & 63.1 & 91.2 & 95.9 & 522.0\\
			VSRN ours & \textbf{Wasserstein~(\ref{eq:wasserstein})} & ellipsoidal & 77.3 & 97.1 & 98.7 & 63.7 & 91.2 & 96.0 & 524.0\\
			\bottomrule
		\end{tabular}
        }
	\end{center}
	\caption{Retrieval ablation study on the MS-COCO validation set. ft stands for fine-tuned. Best results in terms of sum of recalls in each group are highlighted in \textbf{bold}. The numbers in parantheses denote the equation number of the similarity metric in the main manuscript.}
	\label{tab:new_spherical_dev}
\end{table*}

Recall that we use the negative of the KL divergence as one of our similarity metrics.
Table~\ref{tab:new_spherical_dev} contains the recall values for the negative KL divergence, when we take the  caption distribution as reference, and measure the divergence of the image distributions from this reference distribution. In Table~\ref{tab:kl_reverse_dev},
we provide the results (on MS-COCO validation) for the other possibility, where we measure the KL divergence of the caption distributions from the (reference) image distributions. We can see that the sum of recalls (rsum) drops in all cases (with a maximum drop of 2.3\%) when we change the order of the distributions for the KL divergence.


\begin{table*}
	\begin{center}
        \small{
        \begin{tabular}{@{}lllllllll@{}}
			\toprule
			\multirow{2}{*}{model} &\multirow{2}{*}{shape}& \multicolumn{3}{c}{image-to-text} & \multicolumn{3}{c}{text-to-image}&rsum\\
			& &  R@1 & R@5 & R@10 &  R@1 & R@5 & R@10\\
			\midrule
			VSE++ ours & spherical & 52.6 & 85.5 & 93.0 & 43.1 & 76.6 & 87.5 & 438.3 \\
			VSE++ ours & ellipsoidal & 57.5 & 87.6 & 94.0 & 45.6 & 79.6 & 89.9 & 454.3\\
			\midrule
			VSRN ours & spherical & 75.6 & 95.1 & 98.4 & 62.2 & 89.3 & 95.5 & 516.1\\
			VSRN ours & ellipsoidal & 75.5 & 96.7 & 98.8 & 62.4 & 90.7 & 96.3 & 520.4\\
			\bottomrule
		\end{tabular}
        }
	\end{center}
	\caption{Retrieval results on MS-COCO validation set with the negative of KL divergence similarity between the image and caption distributions taken with respect to image representations.}
	\label{tab:kl_reverse_dev}
\end{table*}

\subsection{High and low uncertainty images and captions}

\begin{figure*}
	\begin{center}
	{\includegraphics[width=\textwidth]{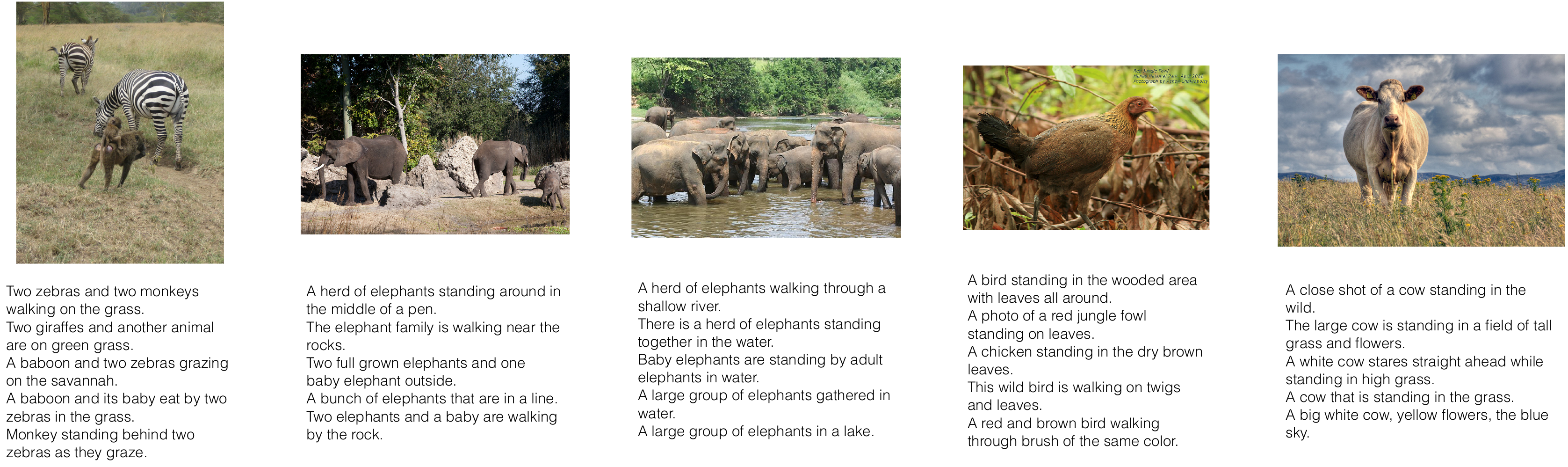}}
	\end{center}
\caption{High uncertainty images and their corresponding plausible matches.}
\label{fig:high_image}
\end{figure*}

\begin{figure*}
	\begin{center}
	{\includegraphics[width=\textwidth]{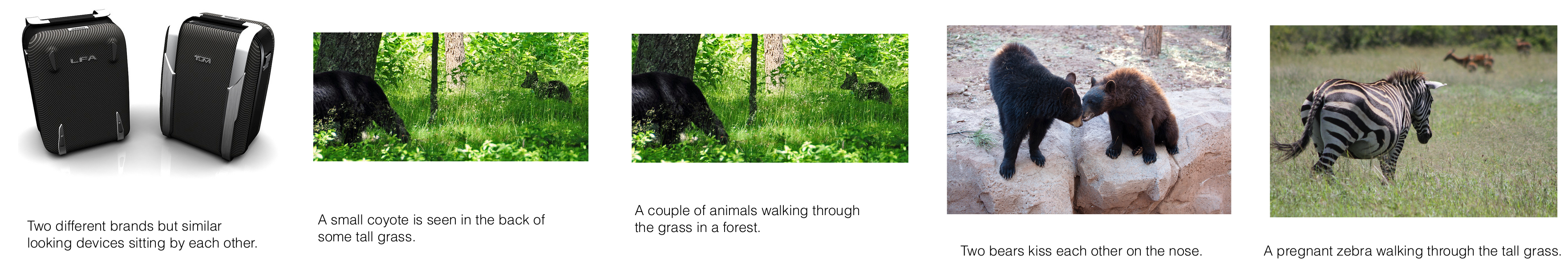}}
	\end{center}
\caption{High uncertainty captions and their corresponding plausible matches.}
\label{fig:high_caption}
\end{figure*}

\begin{figure*}
	\begin{center}
	{\includegraphics[width=\textwidth]{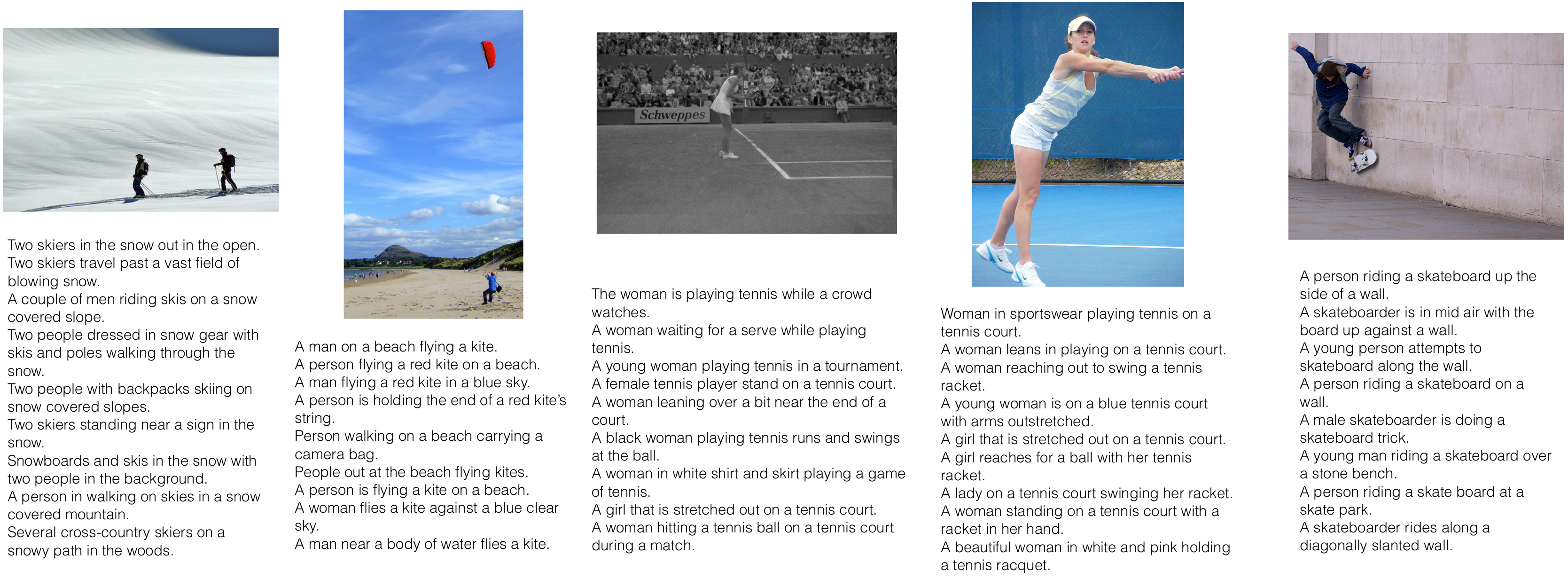}}
	\end{center}
\caption{Low uncertainty images and their corresponding plausible matches.}
\label{fig:low_image}
\end{figure*}

\begin{figure*}
	\begin{center}
	{\includegraphics[width=\textwidth]{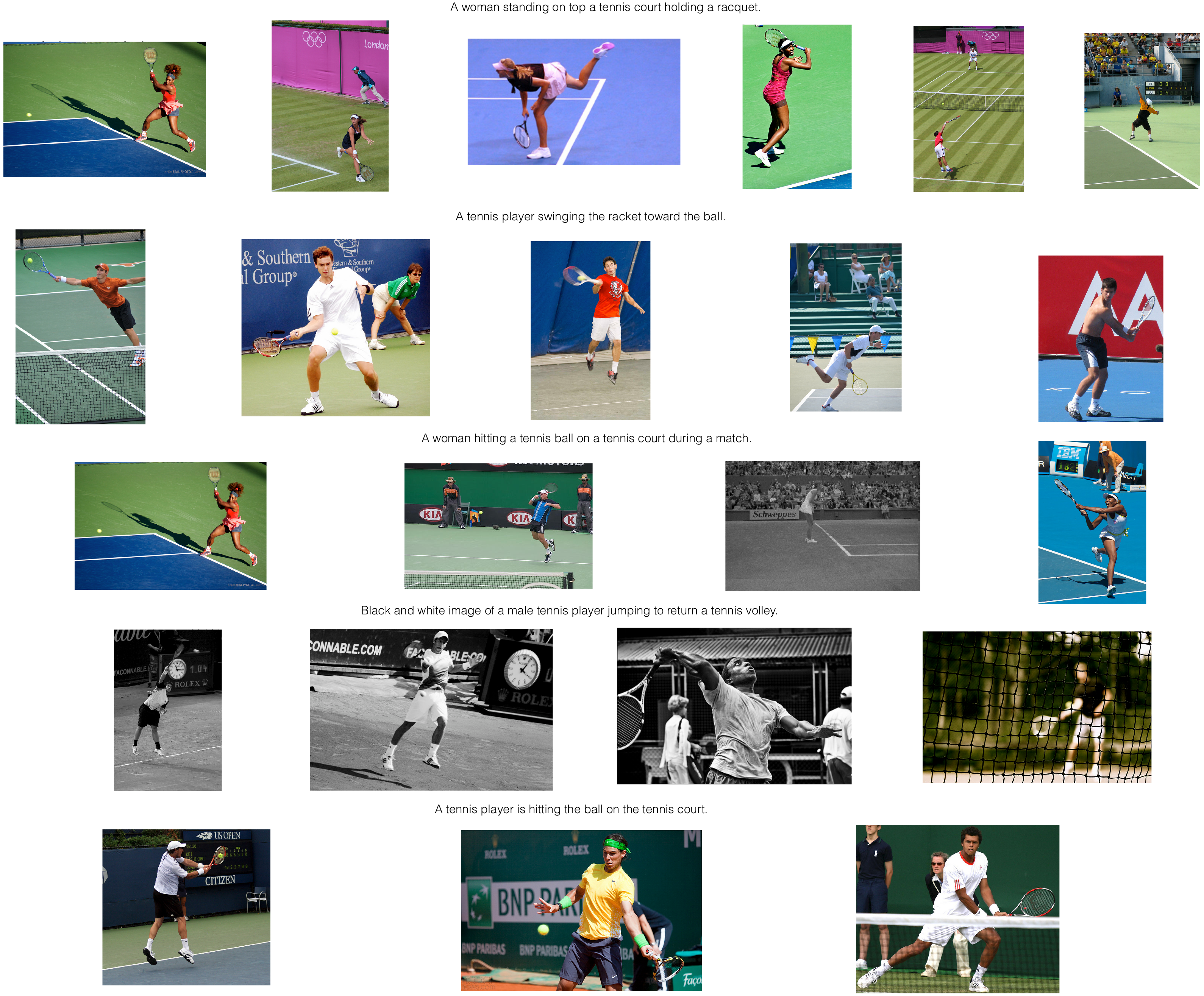}}
	\end{center}
\caption{Low uncertainty captions and their corresponding plausible matches.}
\label{fig:low_caption}
\end{figure*}

In this section, we provide examples with high and low uncertainties along with their plausible matches from the CxC dataset~\cite{Parekh2021}. Figs.~\ref{fig:high_image} and~\ref{fig:high_caption} show five randomly chosen high-uncertainty images and captions, respectively. 
%
Figs.~\ref{fig:low_image} and~\ref{fig:low_caption} show five randomly chosen low-uncertainty images and captions, respectively.

\subsection{Model stability}

Tables~\ref{tab:results_stability_vse} and Table~\ref{tab:results_stability_vsrn} provide stability analyses for the two original models, VSE++ and VSRN, and their probabilistic counterparts (ours), on the test portions of MS-COCO and Flickr30K datasets. Specifically, we generate results for five runs of each model and dataset, and report the mean and standard deviation of the rsum values.
We can see that for VSE++, the probabilistic model shows consistent improvements over the point-based model on both datasets, with a notably larger mean (and small comparable standard deviation) for VSE++ ours compared to VSE++. 
For VSRN, which has a more elaborate image encoding and processing pipeline, we observe a larger variation in performance between runs. Nonetheless, based on these numbers we can still see that the probabilistic VSRN (ours) is competitive (if not better than) the point-based VSRN.

\setlength{\tabcolsep}{7pt}
\begin{table}
	\begin{center}
        \small{
        \begin{tabular}{@{}llll@{}}
			\toprule
			\multirow{2}{*}{model} & \multicolumn{2}{c}{MS-COCO} & \multirow{2}{*}{Flickr30K}\\
			 & \multicolumn{1}{c}{1K}  & \multicolumn{1}{c}{5K} &\\
			\midrule
			VSE++ & \(447.74 \pm 0.51\) & \(310.06 \pm 0.81\) & \(363.42 \pm 2.02\)\\
			VSE++ ours & \(452.70 \pm 0.50\) & \(317.02 \pm 1.07\)& \(372.48\pm 1.86\)\\
			\bottomrule
		\end{tabular}
        }
	\end{center}
	\caption{VSE++ rsum means \(\pm\) standard deviations.}
	\label{tab:results_stability_vse}
\end{table}
\setlength{\tabcolsep}{6pt}



\setlength{\tabcolsep}{7pt}
\begin{table}
	\begin{center}
        \small{
        \begin{tabular}{@{}llll@{}}
			\toprule
			\multirow{2}{*}{model} & \multicolumn{2}{c}{MS-COCO} & \multirow{2}{*}{Flickr30K}\\
			 & \multicolumn{1}{c}{1K}  & \multicolumn{1}{c}{5K} &\\
			\midrule
			VSRN & \(507.68 \pm 0.80\) & \(401.66 \pm 1.40\) & \(466.48 \pm 1.88\) \\
			VSRN ours & \(510.76 \pm 2.02 \) & \(405.02 \pm 2.82\) & \(468.96 \pm 2.64\) \\
			\bottomrule
		\end{tabular}
        }
	\end{center}
	\caption{VSRN rsum means \(\pm\) standard deviations.}
	\label{tab:results_stability_vsrn}
\end{table}
\setlength{\tabcolsep}{6pt}